\relax
\documentclass[letterpaper]{article} 
\usepackage{aaai20}  
\usepackage{times}  
\usepackage{helvet} 
\usepackage{courier}  
\usepackage[hyphens]{url}  
\usepackage{graphicx} 
\usepackage{amsmath} 
\usepackage{amssymb}
\usepackage{multicol} 
\usepackage{multirow} 
\usepackage{subcaption}
\usepackage[T1]{fontenc}
\usepackage{graphicx}  
\title{Adaptive Wasserstein Hourglass for Weakly Supervised Hand Pose Estimation from Monocular RGB}
\author{Yumeng Zhang\textsuperscript{\rm 1}, Li Chen\textsuperscript{\rm 1},Yufeng Liu\textsuperscript{\rm 2}, Junhai Yong\textsuperscript{\rm 1}, Wen Zheng\textsuperscript{\rm 2}\\
\textsuperscript{\rm 1}School of Software, Tsinghua University, Beijing, China\\
\textsuperscript{\rm 2}Y-tech, Kwai, Beijing, China
}

\begin{document}

\maketitle

\begin{abstract}

Insufficient labeled training datasets is one of the bottlenecks of 3D hand pose estimation from monocular RGB images. Synthetic datasets have a large number of images with precise annotations, but the obvious difference with real-world datasets impacts the generalization. Little work has been done to bridge the gap between two domains over their wide difference. In this paper, we propose a domain adaptation method called Adaptive Wasserstein Hourglass (AW Hourglass) for weakly-supervised 3D hand pose estimation, which aims to distinguish the difference and explore the common characteristics (e.g. hand structure) of synthetic and real-world datasets. Learning the common characteristics helps the network focus on pose-related information. The similarity of the characteristics makes it easier to enforce domain-invariant constraints. During training, based on the relation between these common characteristics and 3D pose learned from fully-annotated synthetic datasets, it is beneficial for the network to restore the 3D pose of weakly labeled real-world datasets with the aid of 2D annotations and depth images. While in testing, the network predicts the 3D pose with the input of RGB.

\end{abstract}
 
\section{Introduction}

The broad prospects in the fields of VR/AR applications, gesture recognition and robot control make hand pose estimation always be a research focus in the computer vision. In recent years, monocular RGB image-based 3D hand pose estimation \cite{boukhayma20193d,ge20193d,Iqbal_2018_ECCV,spurr2018cross,Zimmermann_2017_ICCV} has made great progress by means of powerful deep learning methods. Nevertheless, this task remains particularly difficult. Firstly, hand pose is so diverse that there are lots of self-occlusion from single viewpoint. Secondly, 3D estimations from single RGB image is an ill-pose problem due to inherent scale and depth ambiguities. These problems make it difficult to annotate the training data, which is crucial for the performance of deep learning methods.

\begin{figure}[t]
    \begin{center}
    \includegraphics[width=0.45\textwidth]{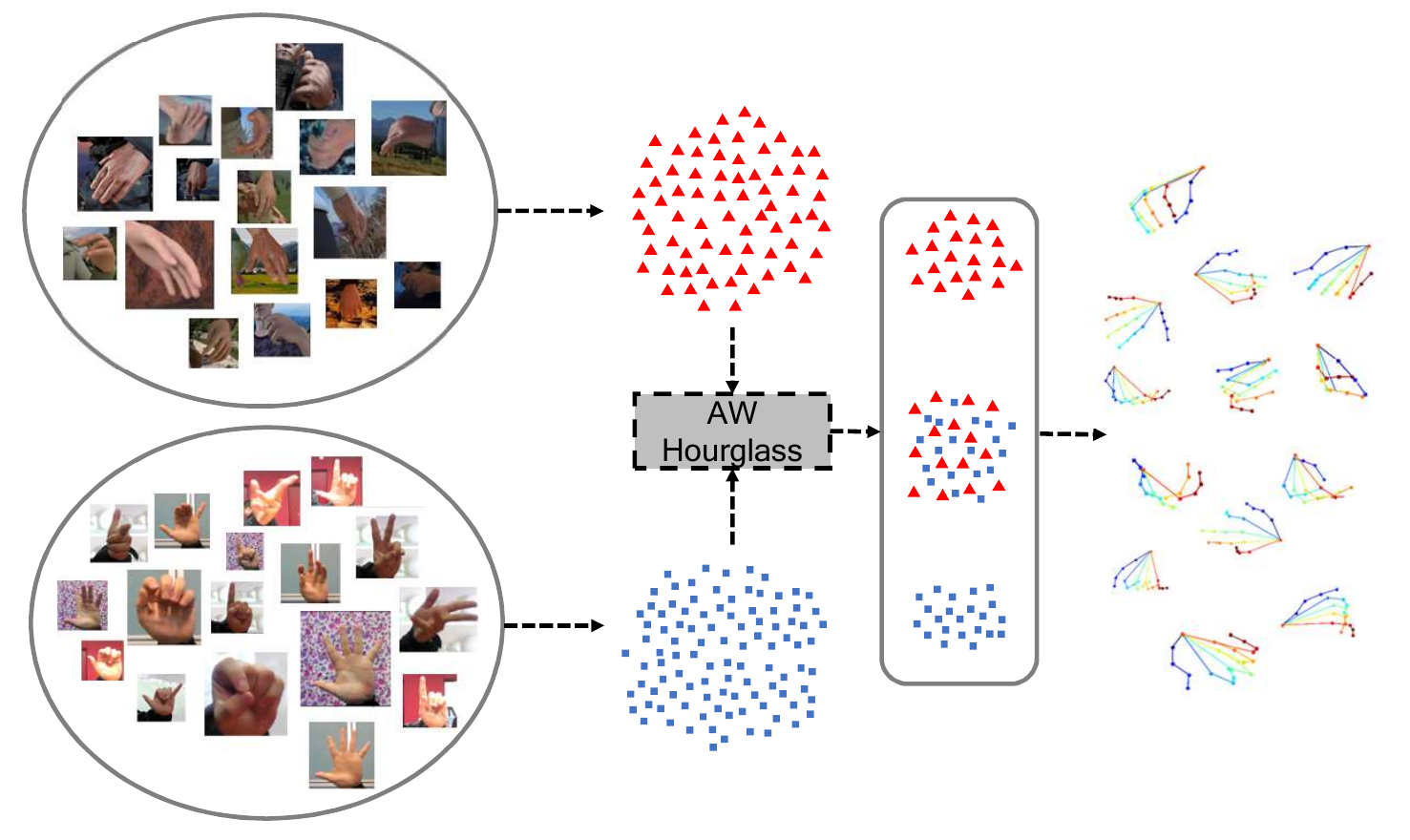}
    \end{center}
    \caption{We propose the Adaptive Wasserstein Hourglass to imporve the performance of weakly-labeled real-world datasets. Our method distinguishes common hand characteristics and dissimilar properties between synthetic and real-world datasets automaticly and bridges the gap between common hand characteristics. As a result, The network begins to notice the common characteristics of hand between two domains, which are beneficial for generalization.}
    \label{fig:briefintroduction}
\end{figure}
                     
Making use of synthetic data \cite{boukhayma20193d,ge20193d,Cai_2018_ECCV,Mueller2018GANerated,Zimmermann_2017_ICCV}, which is of high sufficiency, is the mainstream method to make up the deficiency of training data in hand pose estimation. Zimmermann and Brox \cite{Zimmermann_2017_ICCV} adopted the joint training method of synthetic and real-world datasets, neglecting the difference between the two datasets, which limits the accuracy. Rad et al. \cite{Rad_2018_CVPR} proposed a feature mapping method, which targets at the pose estimation in depth images, to eliminate the gap at the feature level. It requires same pose in synthetic image and corresponding real-world image, which is difficult to apply to pose estimations from monocular RGB images. Mueller et al. \cite{Mueller2018GANerated} tried to narrow the gap by adopting CycleGan \cite{Zhu2017Unpaired} to improve the validity of synthetic images without paired images, but the requirement of only hands images included in real world turns it to be labor intensive. Boukhayma et al. \cite{boukhayma20193d} made use of synthetic datasets to train an encoder to predict the hand and camera parameters so that the came intrinsic parameters are not necessary in the testing phase. However, like all the above-mentioned methods, this method relys on the 3D annotations of real-world datasets, which are more difficult to acquire than camera intrinsics parameters in real-world applications. In order to ease dependence on costly 3D annotations, Cai et al. \cite{Cai_2018_ECCV} adopted 2.5D representation of hand to reduce the uncertainty of 3D pose and utilized synthetic data and the uncostly depth images as the constraints to restore it.  Utilizing a large number of synthetic images for training, Ge et al.\cite{ge20193d} proposed a graph network to simultaneously predict mesh and keypoint position, which works quite well so far without 3D annotations. Utilizing strong constraints to align the two domains is beyond consideration. This leaves room for improvement.

According to our observation, though synthetic and real-world images of hands differ in skin texture and backgrounds, the hand structure, bone length and bone ratio tend to be similar. In order to introduce stronger constraints, we make the network learn the common characteristics, which consist of essential pose-related information. As a result, the 3D predictions of real-world datasets would benefit from learning these characteristics. Domain adaptation methods \cite{ganin2016domain,shen2018wasserstein} are designed to learn the common feature but the big gaps between synthetic and real-world datasets and the high-nonlinear relations between pose and latent features make it difficult for the existing methods to enhance 3D predictions. Therefore, we propose the Adaptive Wasserstein Hourglass, which automatically distinguishes the common hand characteristics and dissimilar properties (e.g. background and texture) so as to bridge the gap of common hand characteristics between the two datasets. Our method finds commonalities between synthetic and real-world database with their differences considered so that the big gap between them can be solved flexibly.  

We conduct several experiments to further improve the network, including the removal of intermediate supervision. In the experiments, we find that the introduction of intermediate supervision makes the features in the Adaptive Wasserstein Hourglass focus more on the annotated information, while neglecting some useful information for real-world 3D pose estimation. Thus the performance gets impacted in the end.

In a word, the main contributions of this paper are as follows: 

\begin{itemize}
\setlength{\itemsep}{0pt}
\setlength{\parsep}{0pt}
\setlength{\parskip}{0pt}

\item We propose an efficient metric to distinguish common hand characteristics and dissimilar
properties between synthetic and real-world datasets. 

\item We propose the Adaptive Wasserstein Hourglass that explores the common characteristics of the synthetic and real-world hands, which can be used as a module in any existing methods with joint training of synthetic and real-world datasets.

\item We study the influence of intermediate supervision in weakly-supervised scenario, which helps us to achieve a further improvement.

\end{itemize}
\vspace{-0.7em}

We conduct several experiments on two Benchmark datasets RHD \cite{Zimmermann_2017_ICCV} and STB \cite{ZhangJCQXY17}. As synthetic datasets are crucial for the accuracy of real-world datasets, it's unfair to compare the results with different synthetic datasets for training. Our method raises the 3D pose estimation accuracy of real-world datasets by a large margin when compared with methods trained for the same datasets. The results of experiment are also competitive with the state-of-the-art \cite{ge20193d} while only about one-tenth of the synthetic images are used. Mesh information is not be used for it requires complex structure to predict for neural networks.

\section{Related Works}
\subsection{Hand Pose Estimation with Synthetic Datasets}

\begin{figure*}
    \begin{center}
    \includegraphics[width=1\textwidth]{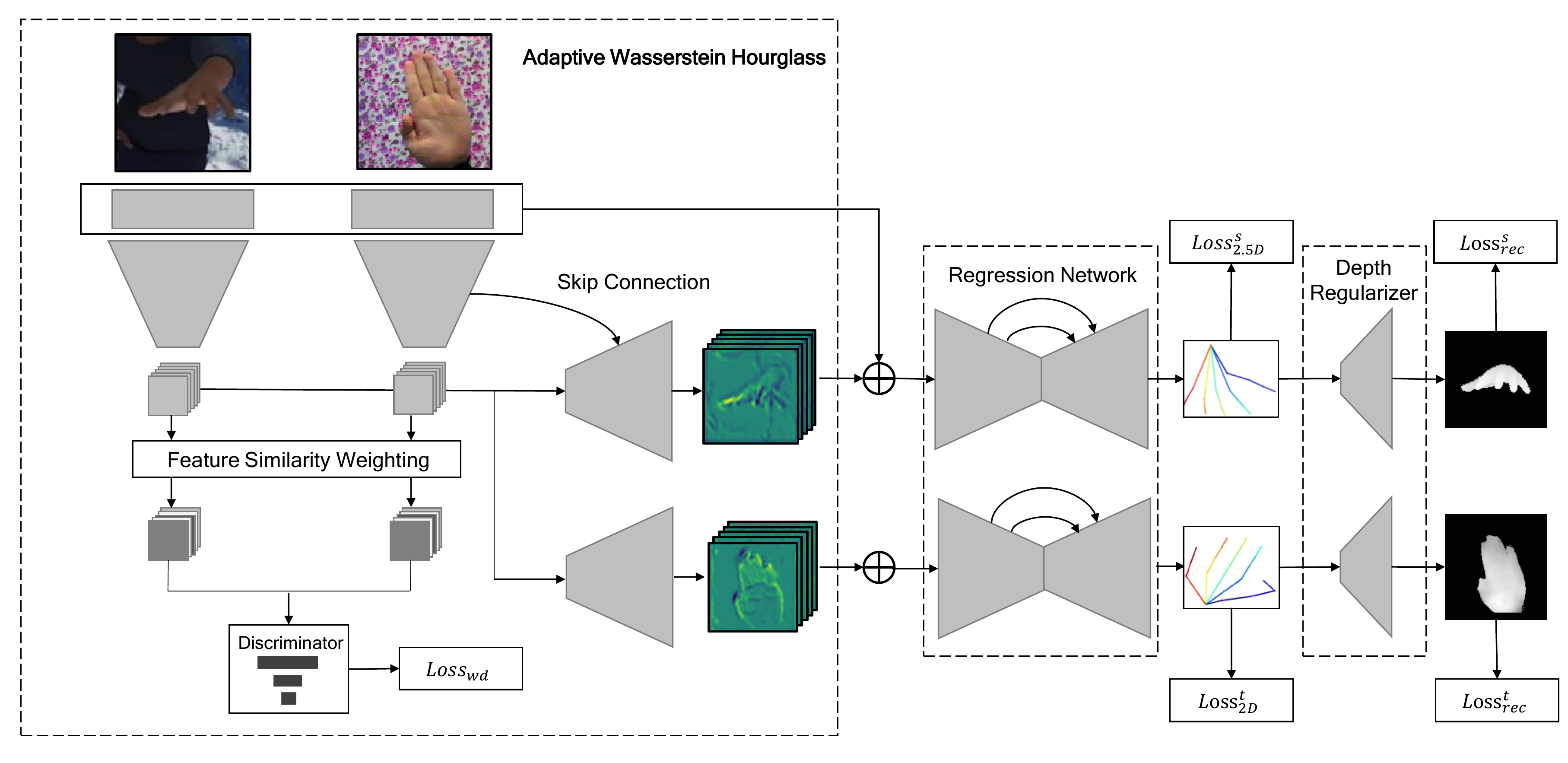}
    \end{center}
    \caption{The overview of our network architecture. The Adaptive Wasserstein Hourglass uses the Feature Similarity Weighting to help network distinguish commonalities of hand and different features such as background. The discriminator is used to brigding the gap between weighted features, which helps network to explore the common characteristics of hand. We add depth regularizer to reduce the depth ambiguities of unlabeled real-world datasets. The subnetworks in the same dashed square mean that they share weights with each other.}
    \label{fig:wholenetwork}
\end{figure*}

With the development of the rendering technology, the validity of computer-generated pictures gets increasingly promoted. Some approaches \cite{Cai_2018_ECCV,Mueller2018GANerated,Zimmermann_2017_ICCV,ge20193d,boukhayma20193d} utilized synthetic images to assist the real-world hand pose estimation. Franziska Mueller et al. \cite{Mueller2017Real} proposed a SynthHands datasets and used it to train hand localization and keypoint regression network. Zimmermann et al. \cite{Zimmermann_2017_ICCV} proposed a rendered hand pose datasets to boost performance of real-world datasets. Boukhayma et al. \cite{boukhayma20193d} adpoted the synthetic datasets to pretrain the encoder to predict the hand and camera parameters and Ge et al. \cite{ge20193d} also proposed a synthetic datasets to train graph network. However, all these methods do not consider the domain gap between synthetic and real-world datasets. Mueller et al. \cite{Mueller2018GANerated} proposed a CycleGan framework to make the synthetic datasets more realistic, but the obvious difference is still in existence. Mahdi et al. \cite{rad2018domain} eliminate the gap between synthetic and real depth images and made the features of RGB images the same as depth images. However, their method takes the depth image for inference and the performance is greatly impacted when using synthetic depth images. Up to now, how to eliminate the domain differences between synthetic datasets and real datasets remains to be a challenge.

\subsection{Domain Adaptation} Domain adaptation techniques \cite{ganin2016domain,Tzeng_2017_CVPR} aim at learning a domain invariant feature for further domain generalization. Domain adaptation methods are rarely applied to hand pose estimation. However, the task bears several resemblances to classification, image segmentation and human pose estimation. Thus, these methods can serve as meaningful reference for designing domain adaptation methods in hand pose estimation. Li et al. \cite{Li_2018_CVPR} proposed a MMD-AAE framework to align the features extracted from multi-domains. Sankaranarayanan et al. \cite{Sankaranarayanan_2018_CVPR} adopted an adversarial training framework for weakly segmentation. However, the relation between images' RGB values and 3D pose are far more nonlinear, so it requires stronger constraints to eliminate the domain gap. Many works in human pose estimation aim to eliminate the gap between datasets collected in laboratory environments and in the wild. Since specified human bone's length should be relatively stable, Zhou et al. \cite{Zhou_2017_ICCV} proposed a GeoLoss to minimize the variance of bone length in a mini-batch as a weak supervision to align two datasets. However, the relation of the two datasets is not established. In contrast, a multi-domain classifier proposed by Yang et al. \cite{Yang_2018_CVPR} views the visual information and relative offsets of the pose as domain prior knowledge to enhance the generalization ability. Although these methods make great achievements in human pose estimation, they are not suitable for hand pose estimation. It is the lacking of realism in synthetic hand rather than difference in background that greatly affects the generalization ability of the model. And some weak constraints failed for the diversity and serious self-occlusion of hand pose. 

\section{Methods}

An overview of our method can be seen in Figure \ref{fig:wholenetwork}. Given a cropped RGB image of a hand $x\in\mathbb{R}^k$, we aim to get 3D pose of the hand, which is represented by the 3D positions of K=21 keypoints in camera coordinate system. In weakly supervised scenario, the study of the relation between latent features and 3D pose almost relys on the synthetic datasets in that only the synthetic datasets have both 2D and 3D annotations. At present, most of methods adopted joint training of two datasets directly and achieved good performance. However, undifferentiated joint training makes the network relies on large capacity to remember the features of the two domains, while ignoring the common hand characteristics of them, which reduces the advantageous effects of synthetic datasets on real-world datasets. In addition, differences of two datasets such as backgrounds and skin texture are domain-specific, which should remain different in the latent space, while the commonalities of hands should be similar as these features are domain-invariant. Therefore, we propose the Adaptive Wasserstein Hourglass to learn the common characteristics by pulling closer features in the latent space selectively. As a result, the knowledge learned from the synthetic datasets can easily transfer to real-world datasets. 

The proposed network architecture consists of three parts: Adaptive Wasserstein Hourglass, regression network and depth regularizer. Adaptive Wasserstein Hourglass extracts the fused features with attention of the common characteristics of two domains and then send the features to the regression network to get the 2.5D coordinates of K keypoints. The depth regularizer is also added to alleviate the depth ambiguities problems. In addition, the regression network and depth regularizer can be replaced by any type of network designed for weakly 3D pose estimation. In this paper, we adopt one of the most advanced network in existing methods.

\subsection{Adaptive Wasserstein Hourglass}

Extracting the features which consist of the common characteristics of two datasets facilitates the 3D pose estimation. However, the Figure \ref{fig:featuremappinga} shows that there are obvious difference between the features extracted by the conventional model \cite{Iqbal_2018_ECCV} with depth regularizer \cite{Cai_2018_ECCV}, from which we conclude that their network has not the ability to recognize the common characteristics of two domains. Therefore, we design the Adaptive Wasserstein Hourglass to make up for this defect.

\subsubsection{Wasserstein Metric}
The wasserstein distance is used to indicate the gap between features of source domain $X_s$ and target domain $X_t$, where the source and target domain represent the synthetic and real-world datasets in our case. Given two continuous distributions $p_r$ and $p_\theta$, with joined distributions $\gamma(p_r, p_\theta)$, the wasserstein distance can be writen as
\begin{center}
    \begin{equation}
    \label{eq:wassersteindistance}
    \begin{split}
    W(p_r,p_\theta) = &\underset{\zeta\in\gamma}{\inf}  \underset{x,y}{\iint}\|x-y\| \zeta(x,y)dxdy\\ 
    &=  \underset{\zeta\in\gamma}{\inf} \mathbb{E}_{x,y~\zeta}[\|x-y\|]
    \end{split}
    \end{equation}
\end{center}

This original definition of wasserstein distance (\ref{eq:wassersteindistance}) is highly intractable \cite{wgan2017}, while the Kantorovich-Rubinstein duality \cite{Villani2009Optimal} demonstrates the distance can be reduced to (\ref{eq:KRwassersteindistance}) under certain conditions, which makes it possible for the neutral networks to estimate this distance.

\begin{center}
    \begin{equation}
    \label{eq:KRwassersteindistance}
    \begin{split}
    W(p_r,p_\theta) = &\underset{\|f\|_L\leq1}{\sup} \mathbb{E}_{s\sim{p_r}}[f(s)] - \mathbb{E}_{s\sim{p_\theta}}[f(t)]
    \end{split}
    \end{equation}
\end{center}

where f is a function with Lipschitz constant of 1.

Wasserstein distance can be used for transfer learning \cite{shen2018wasserstein}. Given a feature representation $z$ by a feature extractor $f_e:\mathbb{R}^k\rightarrow\mathbb{R}^n$. The goal of using wasserstein distance is to make the posterior probability $q_{\theta_e}(z|x_s)$ and $q_{\theta_e}(z|x_t)$ as similar as possible. A discriminator $f_d:\mathbb{R}^n\rightarrow\mathbb{R}$ is adopted to estimate the wasserstein distance by maxmizing the following loss function:

\begin{center}
    \begin{equation}
    \label{eq:loss_wd}
    \begin{split}
    \mathcal{L}_{wd}^*(x^s, x^t) = \mathop{\mathbb{E}}\limits_{x^s \in X^s} [f_d(f_e(x^s))] - \mathop{\mathbb{E}}\limits_{x^t \in X^t}[f_d(f_e(x^t))]
    \end{split}
    \end{equation}
\end{center}

One approach to enforce the Lipschitz constraint is clipping the weights to [-c, c] of the discriminator \cite{wgan2017}. However, the gradient vanishing or exploding problems may be caused by this simply clipping. Thus Gulrajani et al. \cite{WganGPNIPS2017} apply the gradient penalty to promote the convergence of network. The loss function can be writen as 
\begin{center}
    \begin{equation}
    \begin{split}
    \mathcal{L}_{g}(x^s, x^t) = &\mathcal{L}_{wd}(x^s, x^t) + \\
    &\lambda_{gp} \mathop{\mathbb{E}}\limits_{x_m \in X^m}(\|\nabla_{x_m}f(x_m)\|_2-1)^2
    \end{split}
    \label{eq:losswd}
    \end{equation}
\end{center}

where $x^m$ is sampled from $x^s$ and $x^t$ with rate t, which uniformly sampled between 0 and 1.

\subsubsection{Feature Similarity Metric}

There are differences in background and texture between synthetic datasets and real-world datasets, and the background and texture play an important role in determining the area of the hand. The existing domain adaptation methods ignore these differences directly, which results in the confusion of the network and increases the difficulty of the training. In order to solve this problem, we propose an adaptive feature similarity metric to distinguish different features.

Our Adaptive Wasserstein Hourglass takes a batch images of two domains (half-half) as input. Then these images are emmbedded into a latent space z by an encoder $E = f_e:\mathbb{R}^k\rightarrow\mathbb{R}^{C\times H \times W}$, where C represents the number of channels and (H,W) are the size of the features of per channel. The $i_{th}$ channel of features of synthetic datasets are denoted as $z_s^i$, and those of real world datasets are denoted as $z_t^i$. We assume that the features of each channel represent a specific meaning because they are obtained through the same filter. At the same time, features of common hand characteristics should have similar patterns. Thus we define the following metric \ref{eq:featureweight} to calculate the similarity between two datasets channel by channel.  

\begin{center}
    \begin{equation}
    \alpha^i(z_s, z_t) = \sum^{H \times W} \hat{z}_s^i \odot \hat{z}_t^i
    \label{eq:featureweight}
    \end{equation}
\end{center}

$\hat{z}^i$ means that the $i_{th}$ channel of features is normalized. 

We multiply $\hat{z}_s^i$ and $\hat{z}_t^i$ channel by channel after normalization, and then sum them up as the weight of the $i_{th}$ channel of features. The more similar of the features, the higher weights are assigned. Thus in the process of pulling the weighted features of two domains closer to each other, features of common hand characteristics are drawn closer and closer, while features such as different background and texture remain the same.

The Wasserstein distance loss with feature similarity metric is shown in formula \ref{eq:loss_wd_fs}:

\begin{center}
    \begin{equation}
    \label{eq:loss_wd_fs}
    \begin{split}
    \mathcal{L}_{wd}(x^s, x^t) &= \mathop{\mathbb{E}}\limits_{x^s \in X^s} [f_d(\alpha(x_s, x_t) \times f_e(x^s))] \\
    &- \mathop{\mathbb{E}}\limits_{x^t \in X^t}[f_d(\alpha(x_s, x_t) \times f_e(x^t))]
    \end{split}
    \end{equation}
\end{center}

Latent space $z$ learning process adjusts the weights of encoder by back propagation and the decoder utilizes this unified latent space $z$ to find proper representations for regression. As a result, the output of Adaptive Wasserstein Hourglass can be modified. We also add skipping connections to consolidnate features across scales \cite{newell2016stacked}. The detail of the network architecture can be found in supplementary document.

\subsection{Regression Module and Depth Regularizer}

\textbf{Scale Invariant 2.5D Representation}: In order to alleviate the problem of depth ambiguities, we adopt the scale invariant 2.5D representation $P^{2.5D} = \left \{ {P_i^{2.5D}} \right \}_{i=1}^{K}$, already used in \cite{Sun_2017_ICCV,Cai_2018_ECCV,Iqbal_2018_ECCV}, which consists of keypoint 2D locations in pixel coordinates $P_i^{2D}= (x_i^p, y_i^p)$ and normalized root-relative depth value $Z_i^{n}$, which is obtained by subtracting the root coordinates and then normalized by making a specific bone length to a constant C.
\begin{center}
    \begin{equation}
    P_i^n = \frac{C}{d} \cdot(P_i - P_{root})
    \end{equation}
\end{center}
where $P_{root}$ stands for the root coordinates and d = $\|P_{k1} - P_{k2}\|_2$ represents the length of choosen bone between k1 and k2 joints. We follow the same setting in \cite{Cai_2018_ECCV}, using the palm keypoint as the root keypoint. As for selecting the specific bone, we find that the 2D predictions of metacarpals are more stable leading us to choose these bones to normalize the coordinates.

\textbf{Regression Network}: \cite{Iqbal_2018_ECCV} proposed a latent 2.5D heatmap network which achieves best accuracy in fully-supervised method. We adopt similar architecture for 2.5D pose estimation. We employ the L1 loss between predicted 2.5D coordinates $P^{2.5D}$ and the corresponding ground truth $(P^{2.5D})^{gt}$ to avoid using the designed target heatmaps, which is proved to be effective in \cite{Iqbal_2018_ECCV}. The regression loss is 
\begin{center}
    \begin{equation}
    \label{eq:loss_25d}
    \begin{split}
    \mathcal{L}_{2.5D}(P^{2.5D}, (P^{2.5D})^{gt}) = &\sum_{i=1}^{K} (\|P_i^{2D} - (P_i^{2D})^{gt} \|_1 + \\
    &\lambda_{alpha}\|Z_i^{n} - (Z_i^{n})^{gt}\|_1)
    \end{split}
    \end{equation}
\end{center}
where the K represents the number of keypoints.

\textbf{Depth Regularizer}: Depth images offer shape and depth information of the hand, which are good supplements in weakly-supervised scenario. We also adopt depth regularizer module similar to Cai et al. \cite{Cai_2018_ECCV} to limit the range of depth prediction. The network takes 2D ground truth of keypoints and root-relative scale-normalized depth predictions as input and generates the normalized depth image $D_{n}$. 
\begin{center}
    \begin{equation}
    \label{eq:loss_d}
    \begin{split}
    D_{n} = \sum_{i,j} \frac{d_{max}-d_{ij}}{d_{range}}
    \end{split}
    \end{equation}
\end{center}

where the $d_{max}$ and $d_{range}$ represent the max depth value and depth range of hand, and $d_{ij}$ represents the depth value at the location (i,j). 

The loss in this module can be writen as follows:
\begin{center}
    \begin{equation}
    \label{eq:loss_reg}
    \begin{split}
    \mathcal{L}_{reg}(D_{n}, D_{gt}) = \|D_{n} - D_{gt} \|_1
    \end{split}
    \end{equation}
\end{center}

\subsection{Training}

Combining the losses in Eq. (\ref{eq:loss_wd}), (\ref{eq:loss_25d}), (\ref{eq:loss_reg}), the overall loss can be obtained as 
\begin{center}
    \begin{equation}
    \begin{split}
    \mathcal{L} = &-\lambda_{wd}\mathcal{L}_{wd} + \lambda_{res}(\mathcal{L}_{2.5D}^s \\ 
    &+ \mathcal{L}_{2d}^t) + \lambda_{reg}(\mathcal{L}_{reg}^s + \mathcal{L}_{reg}^t)
    \end{split}
    \end{equation}
\end{center}

As the real-world datasets tend to have smaller number of images than synthetic datasets, we replicate some images of real-world datasets to make them equal. In the experiments, we find that pretrain the network on synthetic datasets without discriminator is crucial to achieve a high-quality pose estimation network. After pretraining, the whole network is trained in an end-to-end manner. 

\section{Experiments}

\subsection{Datasets and Metrics}

Our model is trained with two publicly available datasets: Rendered Hand Pose Dataset (RHD) \cite{Zimmermann_2017_ICCV} and Stereo Hand Pose Tracking Benchmark (STB) \cite{ZhangJCQXY17}.

RHD is a synthetic hand pose datasets. It contains 41258 training images and 2728 test images. The training set contains 16 characters performing 31 actions and the test set has 4 characters performing 8 actions. Precise 2D and 3D annotations for 21 keypoints are provided, so are the mask of hands and depth images. There are large variations in viewpoints and hand poses, which can be a good learning reference for real-world datasets.

STB is a real-world datasets. It contains 12 sequences with 6 difference backgrounds. Stereo and depth images were captured from a Point Grey Bumblebee2 stereo camera and an Intel Real Sense F200 activate depth camera simultaneously. 2D and 3D annotations for 21 keypoints are provided. 

End-Point-Error (EPE) and the Area Under the Curve (AUC) on percentage of correct keypoints are used to evaluate the performance of our method. We only report the performance of STB datasets since the synthetic datasets can be viewed as a constraint to restore the 3D pose of STB datasets in weakly supervised scenario. The coordinates relative to the root joint are sufficient in most of applications, thus we assume the global hand scale and the root depth are known in order to report EPE and PCK curve based on 3D hand joint locations in camera coordinate system, following the same condition used in \cite{Zimmermann_2017_ICCV,Cai_2018_ECCV}.

\subsection{Quantitative Results}

\begin{figure}
   \begin{center}
   \includegraphics[width=0.5\textwidth]{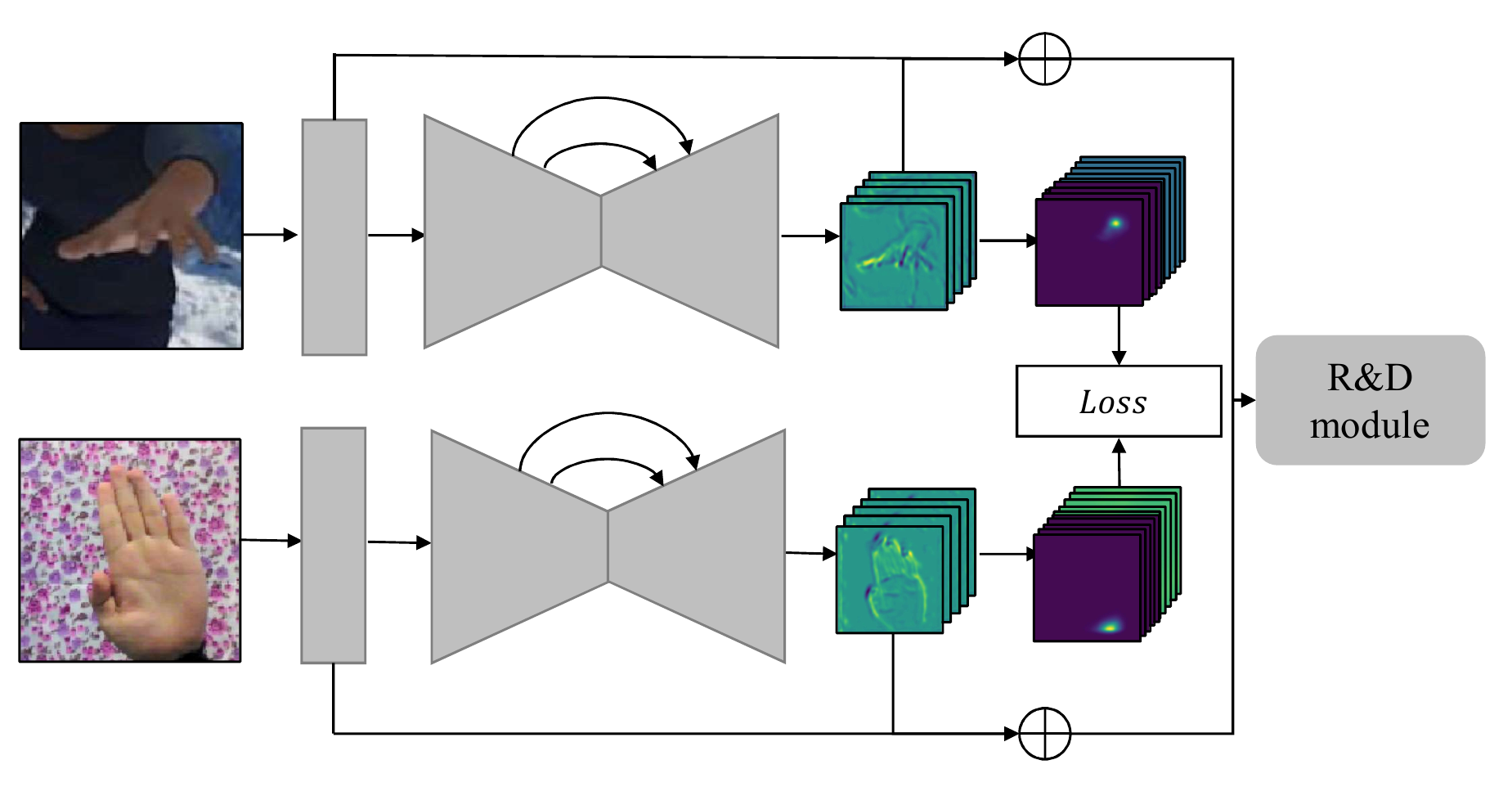}
   \end{center}
   \caption{We add several convolution layers before the 2.5D heatmaps and send these features to the R\&D module (Regression Network and Depth Regularizer). The goal of designing this network is to study the influence of intermediate loss on the heatmaps.}
   \label{fig:FSSvalid}
\end{figure}

\textbf{Rethinking of the Intermediate Supervision}: As there are no 3D annotations of real-world datasets, we think the intermediate supervision would lead the network to focus more on labeled information, neglecting the knowledge necessary for unlabeled 3D pose. To verify our idea, we conduct several experiments to study the influence of intermediate supervision using two-stack network of \cite{Iqbal_2018_ECCV} with depth regularizer module \cite{Cai_2018_ECCV} and the designed network in Figure \ref{fig:FSSvalid}, which adds some convolutional layers before 2.5D heatmaps and sends features in these layers instead of latent 2.5D heatmaps to the next stack. 

\begin{figure}[htbp]
    \centering
    \begin{subfigure}[b]{0.1\textwidth}
    \includegraphics[scale=0.18]{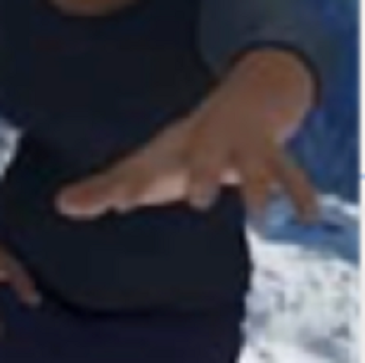}
    \caption{input}
    \end{subfigure}
    \centering
    \begin{subfigure}[b]{0.1\textwidth}
    \includegraphics[scale=0.18]{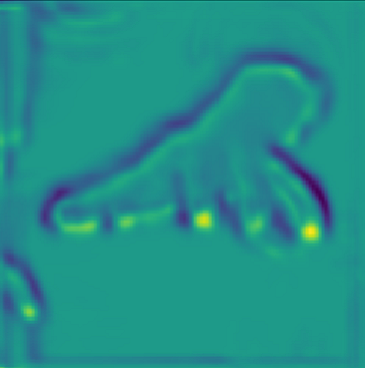}
    \caption{wo/FSS}
    \end{subfigure}
    \centering
    \begin{subfigure}[b]{0.1\textwidth}
    \includegraphics[scale=0.19]{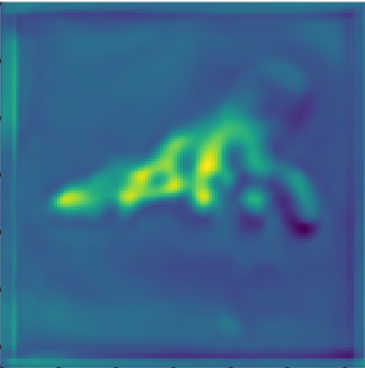}
    \caption{w/FSSC}
    \end{subfigure}
    \centering
    \begin{subfigure}[b]{0.1\textwidth}
    \includegraphics[scale=0.19]{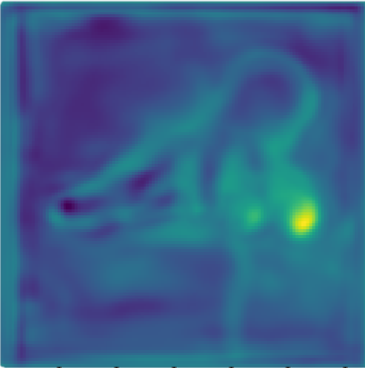}
    \caption{w/FSS}
    \end{subfigure}
    \caption{intermediate heatmaps}
    \label{fig:validonestackloss}
\end{figure}

\begin{figure*}
    \centering
    \begin{subfigure}[b]{0.33\textwidth}
    \includegraphics[scale=0.38]{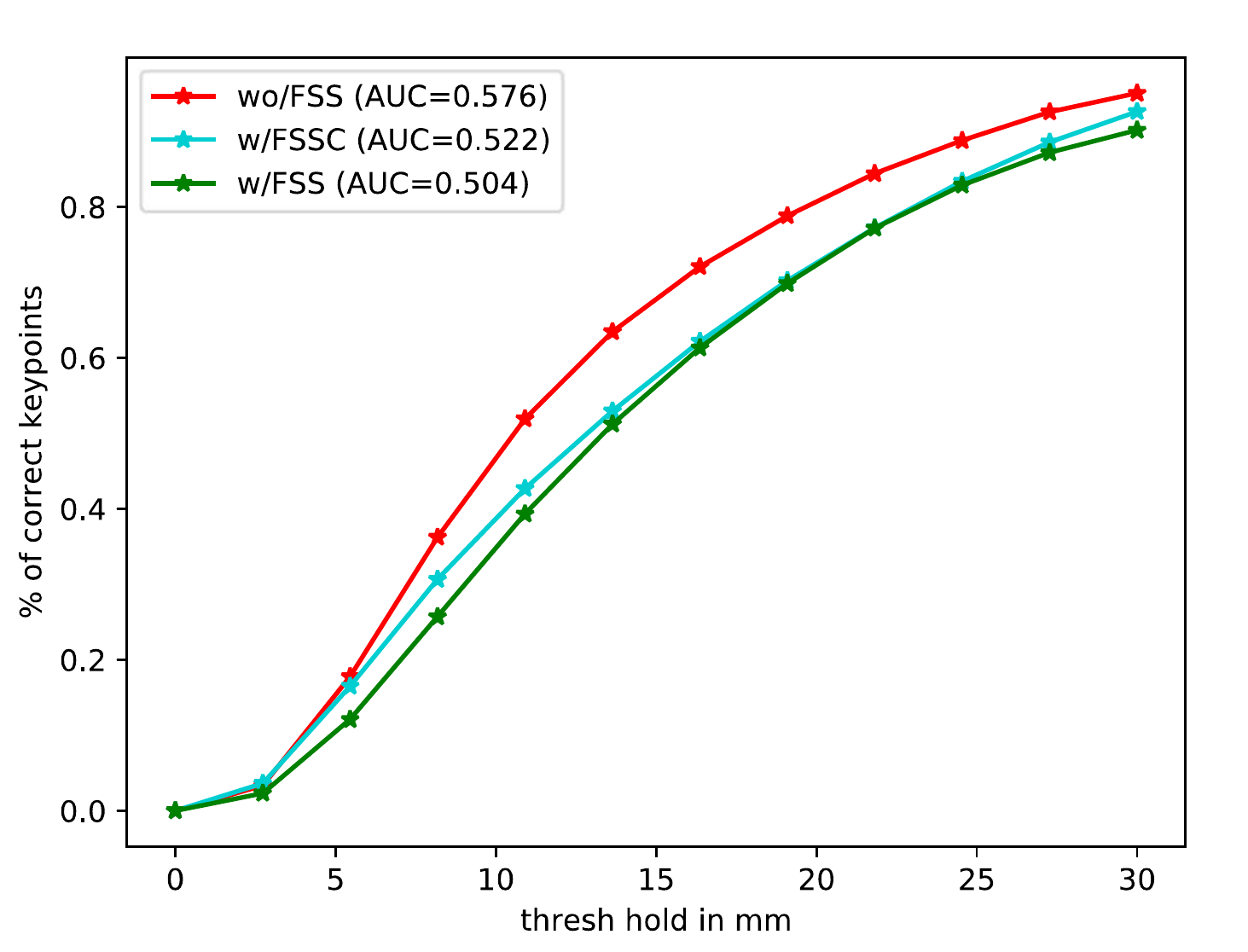} 
    \caption{}
    \label{fig:baselines}
    \end{subfigure}
    \centering
    \begin{subfigure}[b]{0.33\textwidth}
    \includegraphics[scale=0.38]{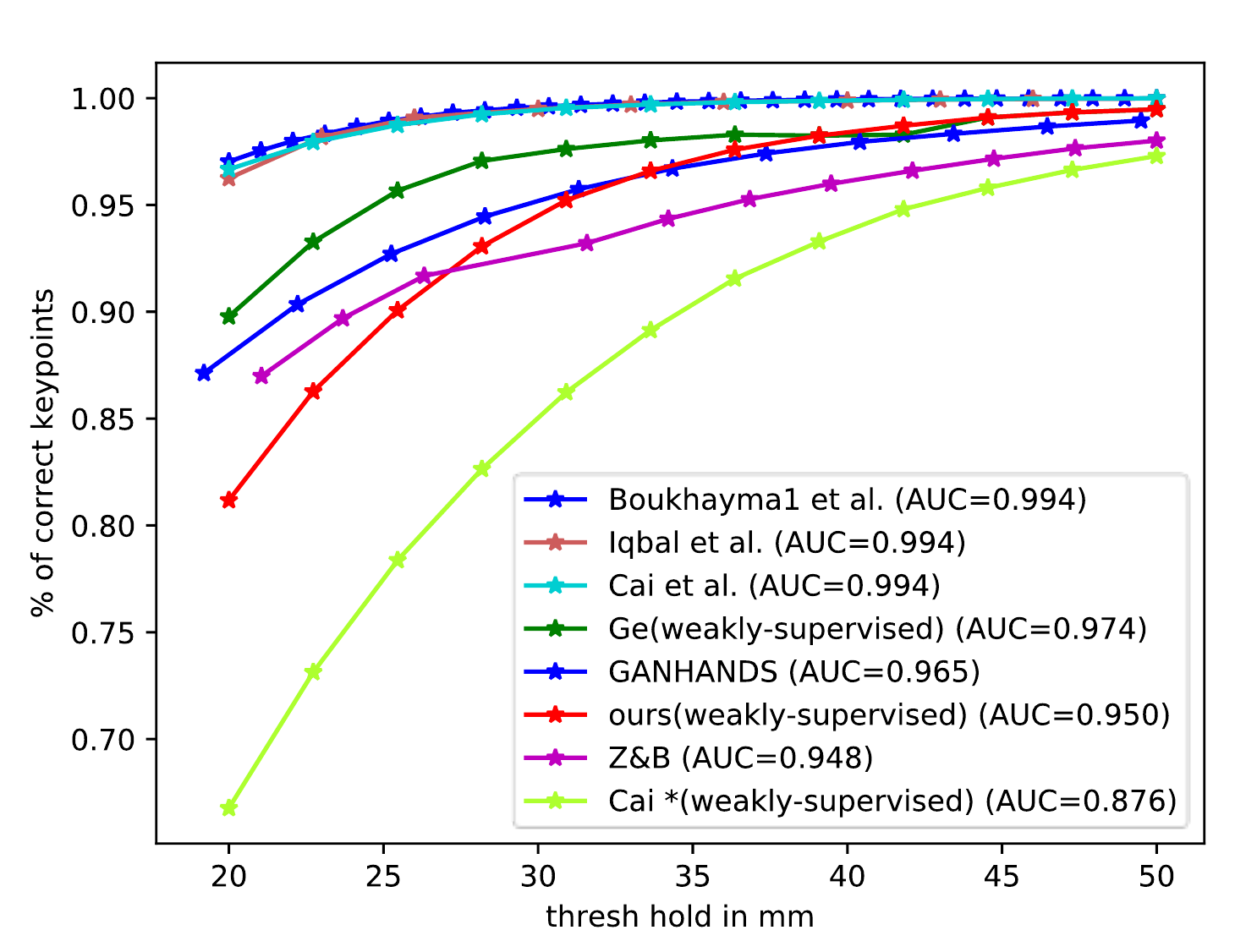}
    \caption{}
    \label{fig:resultsin2050}
    \end{subfigure}
    \centering
    \begin{subfigure}[b]{0.33\textwidth}
    \includegraphics[scale=0.38]{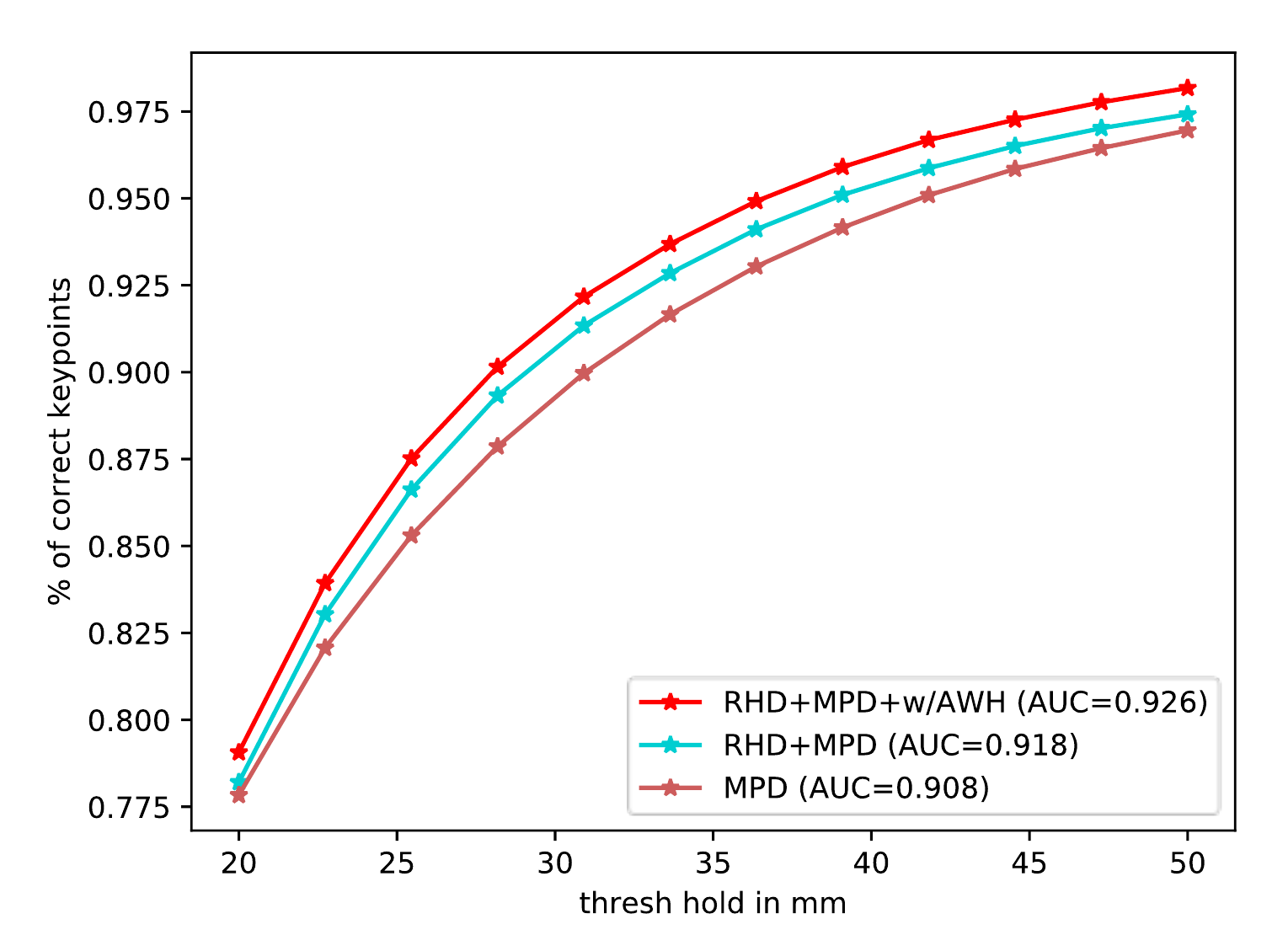}
    \caption{}
    \label{fig:resultsin030}
    \end{subfigure}
\caption{(a) Comparisons with designed baselines to study effects of intermediate supervision. (b) Comparisons with state-of-the-art methods. Our method outperforms some fully-supervised methods. (c) Experimental results on MPD \cite{gomezdonoso2019large-scale} datasets.}
\label{fig:pcks}
\end{figure*}

The supervision at the first stack of \cite{Iqbal_2018_ECCV} is denoted as FSS (First Stack Supervision) while supervision at the first stack of the designed network is denoted as FSSC (First Stack Supervision at additional Convolutional layer). To evaluate the performance of our method on Stereo datasets, we follow the same evaluation protocal used in \cite{Cai_2018_ECCV}. We train the model with 10 sequences and test on the other 2 sequences. 

As the experimental results shown in the Figure \ref{fig:baselines}, it can be observed that adding first stack supervision affects the accuracy greatly. In order to explore the reasons for the variations in accuracy, we visualize the first-stack heatmaps of w/FSS, w/FSSC and wo/FSS, as shown in Figure \ref{fig:validonestackloss}. The heatmaps of wo/FSS include rich information of the image while the other two mainly focus on a local region. Recently some meta learning methods \cite{bauer2017discriminative,oreshkin2018tadam} pointed out that a good feature extractor are crucial for generalization. Weakly supervised hand pose estimation can also be viewed as a generalization from synthetic datasets to real world datasets. Adding intermediate supervision makes the features so discriminative that some important information get lost, which impacts the performance.  

In addition, we conduct experiments to study the influence of different losses in intermediate supervision. We denote 2D and 3D supervision of RHD \cite{Zimmermann_2017_ICCV} as $R_{2D}$ and $R_{3D}$, while the 2D supervision of STB \cite{ZhangJCQXY17} are denoted as $S_{2D}$. We add different losses in the first stack of \cite{Iqbal_2018_ECCV} according to permutation and combination and the results are shown in Table \ref{tab:iL}.

\renewcommand{\arraystretch}{1.5}
\begin{table}[tp]
  \centering
  \small
  \fontsize{9}{8}\selectfont
    \begin{tabular}{|c|c|c|c|c|c|c|}
    \hline
    \multirow{1}{*}{Experiments}&3D Mean EPE(mm)&AUC@30mm\cr
    \hline
    \hline
    R2D&15.7&0.512\cr\hline
    R3D&16.6&0.477\cr\hline
    S2D&14.1&0.553\cr\hline
    R2D+S2D&17.1&0.482\cr\hline
    R2D+R3D&15.9&0.499\cr\hline
    S2D+R3D&16.2&0.486\cr\hline
    R2D+S2D+R3D&15.9&0.504\cr\hline
    wo/IL&{\bf13.1}&{\bf0.576}\cr\hline
    \end{tabular}
    \caption{The left column represents the losses added as intermediate supervision. wo/IL means that we don't use intermediate supervision.}
    \label{tab:iL}
\end{table}

From these experimental results, we conclude that the synthetic datasets annotations in the first stack affect the final performance of keypoint detection. 

\begin{figure}
    \centering
    \begin{subfigure}[b]{0.19\textwidth}
        \includegraphics[scale=0.27]{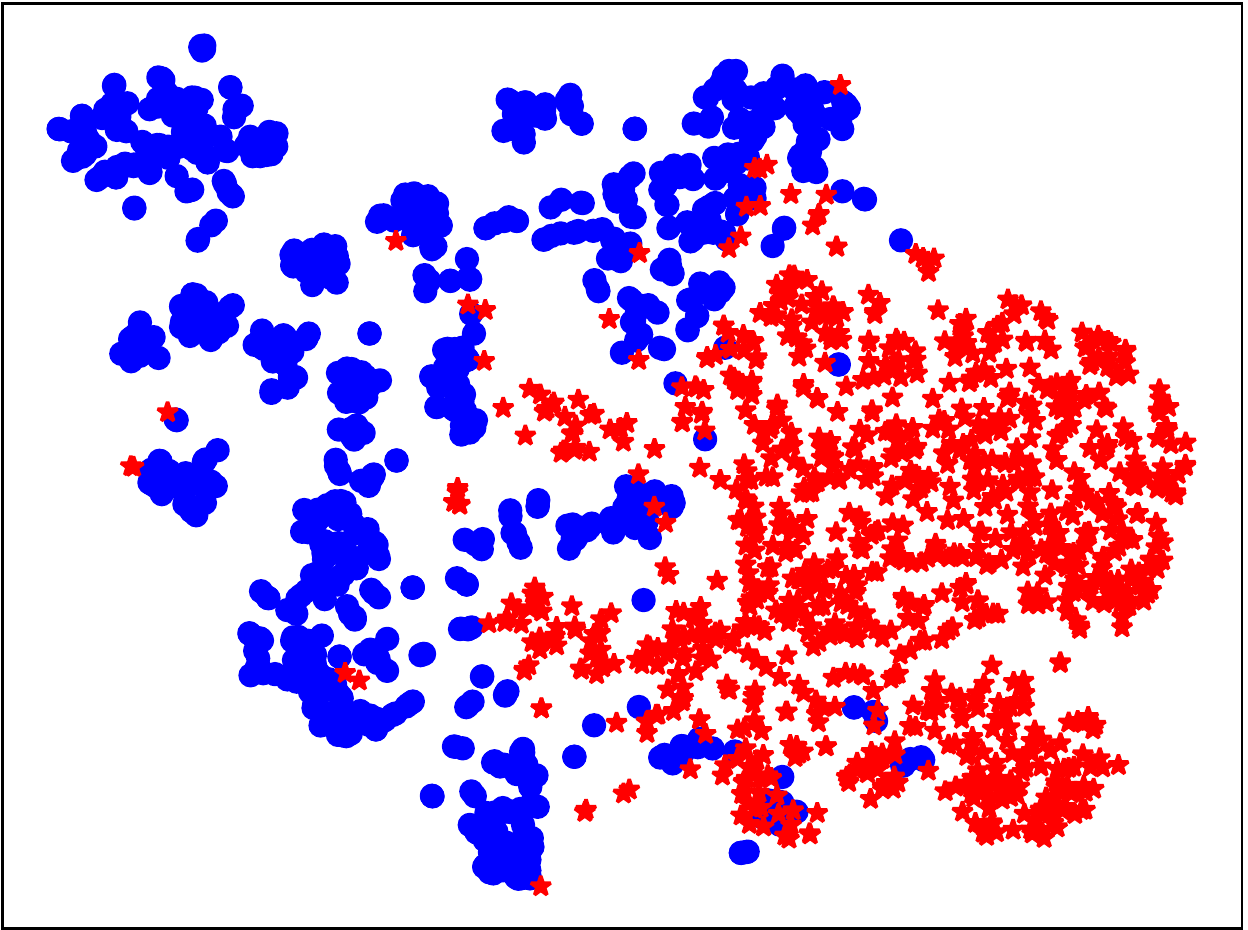}
        \caption{without AWH}
        \label{fig:featuremappinga}
    \end{subfigure}
    \centering
    \begin{subfigure}[b]{0.22\textwidth}
        \includegraphics[scale=0.27]{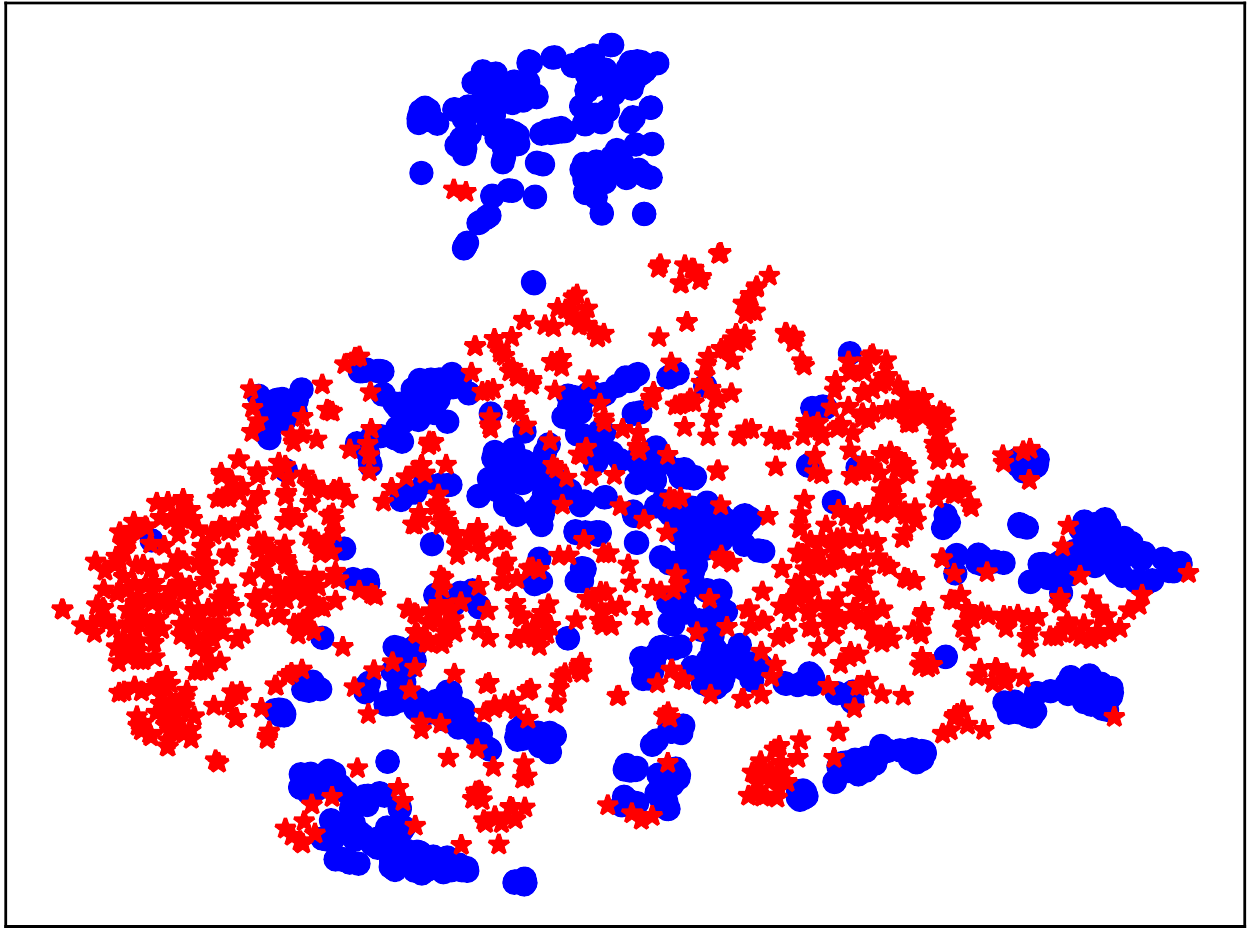}
        \caption{with AWH}
        \label{fig:featuremappingb}
    \end{subfigure}
    \caption{We randomly select 2000 samples of synthetic images and real-world images(half-half) and visualize their features by t-SNE \cite{maaten2008visualizing}. The red points represent samples in source domain while the blue points represent samples the target domain.}
    \label{fig:featuremapping}
\end{figure}

\textbf{Baseline}: To demonstrate the effectiveness of our Adaptive Wasserstein Hourglass, we compare the results with two baselines. For easier representation, the Adaptive Wasserstein Hourglass is denoted as AWH. Then the proposed method can be represented as w/AWH+wo/FSS. Together with the previous experiments, the two baselines are: a) wo/AWH+w/FSS: two-stack network in \cite{Iqbal_2018_ECCV} with depth regularizer module. b) w/WH+wo/FSS: adding a domain adaptation module proposed by \cite{shen2018wasserstein}. 

\begin{figure}[h]
    \begin{center}
    \includegraphics[width=0.45\textwidth]{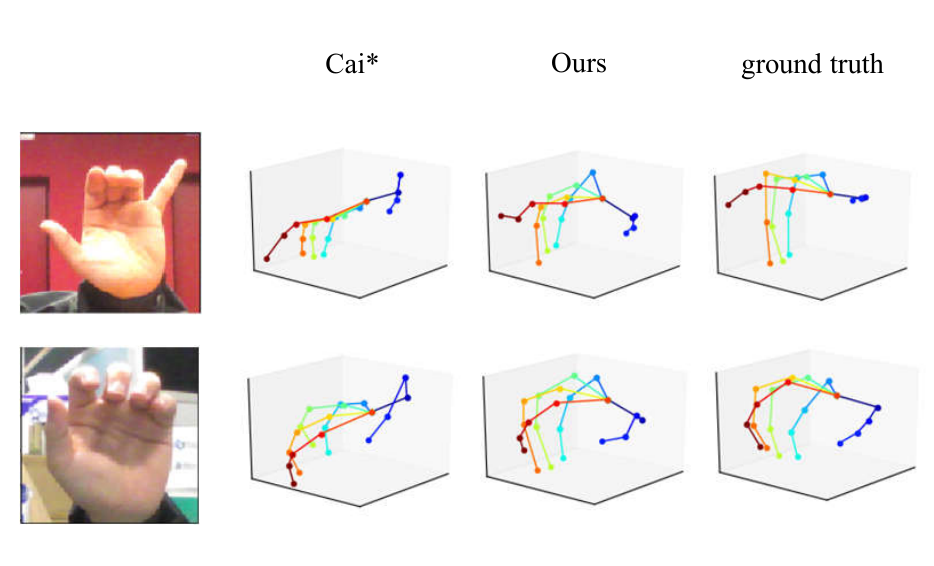}
    \end{center}
    \caption{We compare our results with Cai et al \cite{Cai_2018_ECCV}. Our method predicts 3D pose more accurately and it even correctly restores the 3D pose with severe self occlusion.}
    \label{fig:qualitativeresults_weak}
\end{figure}

The results of these experiments are shown in Table \ref{tab:performance_comparison}. Although domain adaptation methods achieve good performance in classification, they are not suitable for 3D hand pose estimation, where two domains are quite different. We also test methods of \cite{Tzeng_2017_CVPR,ganin2016domain}, but the network becomes extremely difficult to train. Based on above observations, we argue that there are some differences between two domains that can not be ignored. Thus we should distinguish the features, narrow the gap between the features of common hand characteristics while retaining difference of the features related to the background. This is the goal of our Adaptive Wasserstein Hourglass.

\textbf{Comparisons with State-of-the-arts}: 

We compare our method with state-of-the-art fully-supervised and weakly-supervised methods. The comparisons are shown in Figure \ref{fig:resultsin2050}. The methods with * represents the weakly supervised method which adopted same synthetic datasets as us. We achieve competitive results with some fully-supervised methods. Although the performance is slightly lower than Ge's \cite{ge20193d} method, we only use a tenth number of synthetic images without mesh annotations and our approach is more expansive, which can be served as module to be added to existing methods. 

\renewcommand{\arraystretch}{1.5}
\begin{table}[tp] 
  \centering
  \small
  \fontsize{9}{8}\selectfont
  
    \begin{tabular}{|c|c|c|c|c|c|c|}
    \hline
    \multirow{1}{*}{Experiments}&3D Mean EPE(mm)&AUC@30mm\cr
    \hline
    \hline
    wo/AWH+wo/FSS&13.1&0.576\cr\hline
    w/WH+wo/FSS&13.1&0.578\cr\hline
    w/AWH+wo/FSS&{\bf12.8}&{\bf0.589}\cr\hline
    \end{tabular}
    \caption{Comparisons of 3D PCK results of five baselines with the proposed method}
    \label{tab:performance_comparison}
\end{table}
  
\subsection{Qualitative results}

Several visual results of the proposed method are shown in Figure \ref{fig:qualitativeresults_weak}. We compare the results with Cai et al. \cite{Cai_2018_ECCV}, which adopted same training datasets and achieved second best results in existing weakly supervised methods. It can be seen that our model restores these hand pose with self occlusion correctly. More viusal results can be found in supplementary document.  

\textbf{Feature Visualization}: In order to investigate effectiveness of the latent features learned by our Wasserstein Hourglass, we use t-SNE \cite{maaten2008visualizing} to visualize these features extracted by two models--w/AWH+wo/FSS and wo/AWH+wo/FSS. A comparison between Figure \ref{fig:featuremappinga} and Figure \ref{fig:featuremappingb} reveals that Adaptive Wasserstein Hourglass has the ability to minimize the domain gap between similar features as well as keep the distance of domain-specific features.

\subsection{Fully Supervised Scenario}

Although our method is designed for weakly supervised scenario, we find it is still effective in fully supervised situation. We use the MPD \cite{gomezdonoso2019large-scale} real-world dataset, which have 82176 annotated samples, to test our method. Three experiments are done: a) solely training on MPD datasets(MPD). b)joint training with RHD (MPD+RHD). c) joint training with AWH module (RHD+MPD+w/AWH). The results are shown in Figure \ref{fig:resultsin030}. The PCK value has increased form 91.78\% to 92.60\% after adding the AWH module, which means that learning the common characteristics of hand helps improve the utilization of synthetic datasets.

\section{Conclusion}

We present a domain adaptation method for 3D hand pose estimation. We use the idea of meta learning for reference, and propose an adaptive feature extractor, which can extract common characteristics of hands in different domains while maintaining domain-specific features. Our method can effectively improve the accuracy, so as to be used as a module in other methods with good expansibility. In the future, we will try to combine the realization of synthetic datasets with domain adaptation methods to improve the utilization of synthetic datasets.


\bibliography{egbib}
\bibliographystyle{aaai}

\end{document}